\documentclass[letterpaper, 10 pt, conference]{ieeeconf}  % Comment this line out if you need a4paper
\usepackage[utf8]{inputenc}
\usepackage{amsmath, amsfonts}
\usepackage{graphicx}
\usepackage{xcolor}
\usepackage{wrapfig}
\usepackage{adjustbox}
\usepackage{xspace}
\usepackage{epstopdf}
\usepackage{xcolor}
\usepackage{float}
\usepackage{soul}
\usepackage{hyperref}
\usepackage{siunitx}
\usepackage{algorithm}
\usepackage{algcompatible}
\usepackage[capitalize]{cleveref}
\usepackage{booktabs}
\usepackage{multirow}
\usepackage[para]{threeparttable} 
\usepackage{array}
\usepackage{colortbl}
\usepackage{siunitx}

\usepackage{eqexpl}
\eqexplSetDelim{=}

\usepackage{algpseudocode}
\usepackage{array}
\usepackage{tabularx}

\usepackage{wrapfig}
\usepackage[small]{caption}
\usepackage{subcaption}
\usepackage[noadjust]{cite}
\usepackage{comment}

\IEEEoverridecommandlockouts % <- needed for \thanks command

\newcommand{\systemname}{\textit{}\xspace}

\newcommand{\specialcell}[2][c]{%
  \begin{tabular}[#1]{@{}l@{}}#2\end{tabular}}

\title{\systemname \textsl{GenTact Toolbox}: A Computational Design Pipeline to Procedurally Generate Context-Driven 3D Printed Whole-Body Artificial Skins}

\author{Carson Kohlbrenner, Caleb Escobedo, S. Sandra Bae, Alexander Dickhans, and Alessandro Roncone 
\thanks{This work was accepted at the IEEE International Conference on
 Robotics and Automation (ICRA) 2025. Copyright may be transferred
 without notice, after which this version may no longer be accessible. All authors are with the University of Colorado Boulder, 1111 Engineering Drive, Boulder, CO USA. This work is partially supported by NSF FW-
HTF-R grant \#2222952. This work was authored (in part) by the National Renewable Energy Laboratory, operated by the Alliance for Sustainable Energy, LLC, for the US Department of Energy (DOE) under contract no. DE-AC36-08GO28308.
{\tt\small name.surname@colorado.edu}.}
}
\begin{document}
\maketitle

\begin{abstract}

Developing whole-body tactile skins for robots remains a challenging task, as existing solutions often prioritize modular, one-size-fits-all designs, which, while versatile, fail to account for the robot’s specific shape and the unique demands of its operational context. In this work, we introduce \textsl{GenTact Toolbox}, a computational pipeline for creating versatile whole-body tactile skins tailored to both robot shape and application domain. Our method includes procedural mesh generation for conforming to a robot's topology, task-driven simulation to refine sensor distribution, and multi-material 3D printing for shape-agnostic fabrication. We validate our approach by creating and deploying six capacitive sensing skins on a Franka Research 3 robot arm in a human-robot interaction scenario. This work represents a shift from ``one-size-fits-all'' tactile sensors toward context-driven, highly adaptable designs that can be customized for a wide range of robotic systems and applications. The project website is available at \url{https://hiro-group.ronc.one/gentacttoolbox}

\end{abstract}

\section{Introduction}

Whole-body tactile feedback assists humans in effectively interacting with the environment, developing social connections, and avoiding danger \cite{ross2024emergence, hoffmann2022body}. Similarly, robots equipped with whole-body tactile sensors benefit from touch feedback in exploration, interaction, and task execution in unstructured environments \cite{albini2021exploiting, block2021six, roncone2016peripersonal}. To this end, there has been a significant body of work in developing whole-body tactile skins, i.e. arrays of sensors designed to give a robot the sense of touch over the full surface of its body \cite{Dahiya2019LargeArea, Dahiya2019Humanoids, cheng2019comprehensive}. However, state-of-the-art methods are one-off solutions with limited adaptability, and, to the best of our knowledge, no platform currently exists that enables roboticists to design or customize purpose-built, whole-body skins tailored to their specific robot or use case. Interestingly, a robot’s operational context and shape \textsl{can directly influence 
critical design parameters} of a whole-body artificial skin, such as sensing resolution, surface coverage, the skin's softness/compliance, or the bandwidth at which data needs to be processed to successfully complete the task.
% the optimal sensing resolution and surface coverage needed for a tactile skin, reflecting how and where tactile data should be collected. 
For example, precise manipulation in clutter (e.g. \cite{murali2022active, jain2013manipulation, albini2021exploiting, rozlivek2025}) requires high-resolution data to localize force as opposed to a binary and low resolution detection of contact needed for safe physical human-robot interaction (pHRI) (e.g. \cite{erickson2019multidimensional, fan2022enabling, nguyen2018, escobedo2021contact}). 
% \ar{I think you should remove these two paragraphs below altogether. You can't introduce your work twice, it feels weird. So you first show all your motivations and then present your work (which is what you had). Feel free to remove this when you read it.}
% Designing skins that support configurable sensor resolutions is crucial for whole-body tactile solutions that are generalizable to any given robot and operational context. 
% In this work, we seek to bridge this gap and to address the design challenges preventing whole-body tactile sensors from being adopted to a wide range of robots and tasks. In particular, we introduce a design method that allows roboticists to create and tailored whole-body tactile sensors to their specific robots and operational contexts.

% \end{figure}
\begin{figure}
    \centering
    \includegraphics[width=0.47\textwidth]{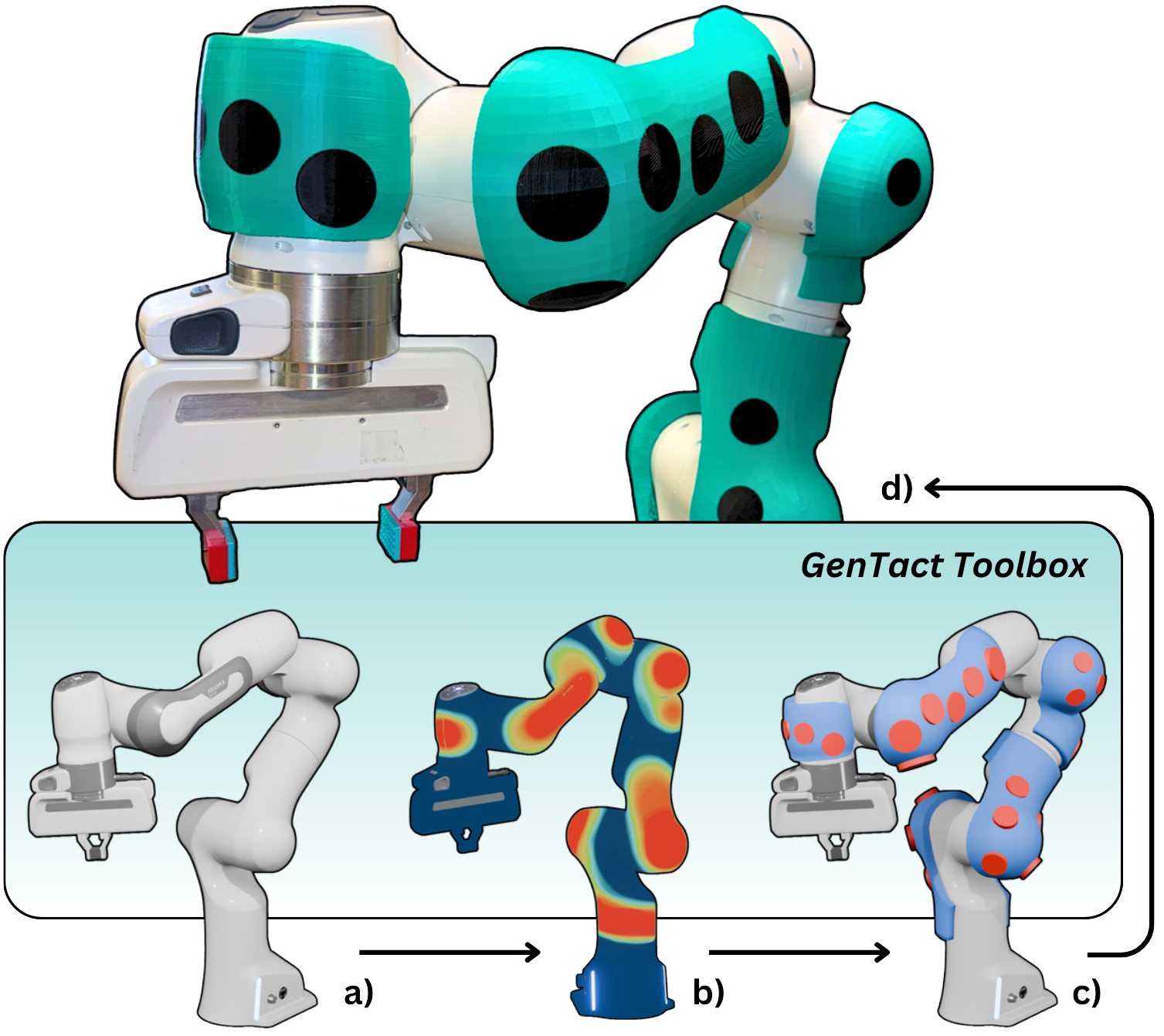}
    \caption{The computational pipeline presented in this work, \textsl{GenTact Toolbox}, generates form-fitting and adaptable whole-body tactile sensors. \textsl{GenTact Toolbox} uses the 3D model of a given robot (a) and a user-generated heat map (b) to create digital meshes of sensor arrays (b) that can be 3D printed as functional tactile sensors (c).}
    \label{fig:cover_pic}
\end{figure}
\begin{figure*}
  \vspace{1.5em}
  \centering
  \includegraphics[width=1\linewidth]{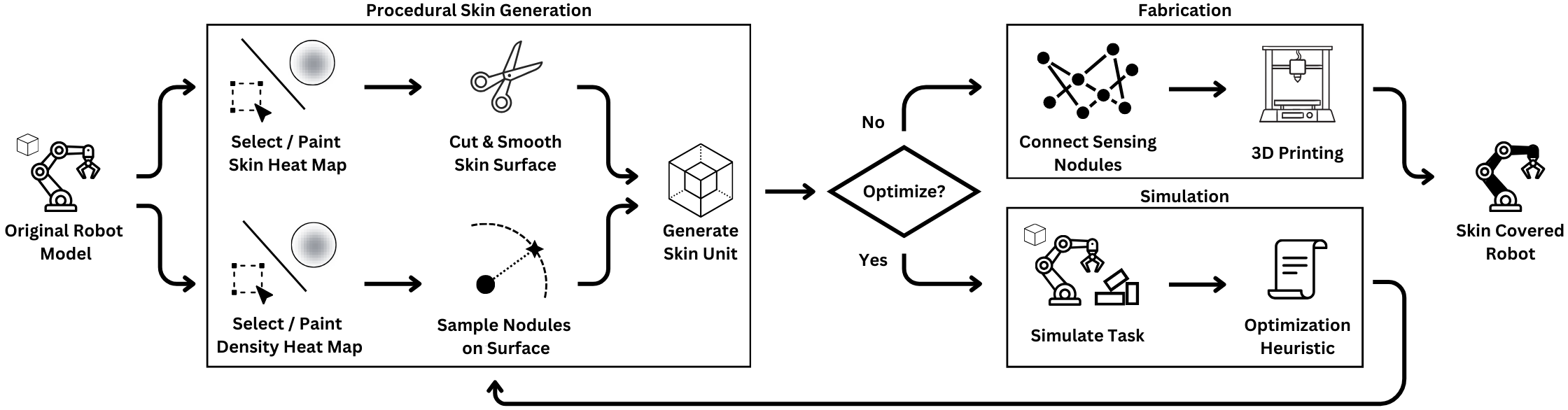}
  \caption{The \textsl{GenTact} pipeline for designing form-fitting and adaptable tactile skins is composed of three stages: procedural generation, simulation, and fabrication. The \textsl{procedural generation} stage (left) generates an initial distribution of sensors that are then passed into the \textsl{simulation} stage (bottom right) to be evaluated and improved based on the task they are used for. Finally, the sensors are connected internally in the \textsl{fabrication} stage (top right) to be printed and deployed on the real robot.}
  \label{fig:flowchart}
\end{figure*}

Another gap this work seeks to address is that current state of the art systems have a high barrier to entry for end users.
The literature focused on designing sensors that conform to robot geometries with modular or flexible designs that wrap over a robot surface \cite{cheng2019comprehensive, maiolino2013flexible, zhou2023tacsuit, teyssier2021human, xu2024cushsense, jiang2024hierarchical}. While wrapping sensors to conform to robot geometry affords generalization after fabrication, it can be problematic at the hardware and software integration levels because it requires manual assembly and localization of each sensor \cite{watanabe2021self}. Of note, robots come with a 3D model that contains precise link dimensions and geometry that can alleviate these integration issues if applied across both the pre-fabrication and post-fabrication stages. It is therefore possible to leverage this 3D model to inform the design of \textsl{a new class of tactile sensors that are generated for a specific embodiment and towards a specific application}. 
% are defined by a set of simple geometric rules. 
This process, inspired by procedural generation techniques used in computer graphics and video games, can be further combined with generalized fabrication processes such as 3D printing.
This in turn allows us to bring the flexibility of procedural generation from the design stage to the manufacturing itself, and enables customizable, form-fitting, functional, and rapidly deployable tactile skins.

% Describe technical approach to solving the problem
In this work, we present \textsl{GenTact Toolbox}, a computational design pipeline that produces \textsl{form-fitting} and \textsl{context-driven} tactile sensors tailored to the unique demands of individual robots and application domains (\cref{fig:cover_pic}). \textsl{GenTact Toolbox} has three main components: a procedural mesh generation stage that algorithmically designs form-fitting and functional tactile sensing skins, a task-driven simulation stage that optimizes sensor placement for a given application, and a multi-material 3D printing phase to fabricate generated skins as capacitive sensors. The first stage of our pipeline has designers indicate the regions of skin coverage and sensor placement over a robot's 3D model with virtual heat maps. The designer can then vary the skin thickness, sensor size, and sensor density within these heat maps by adjusting configurable scalar parameters. In the second stage, the generated skin sensors are simulated in Isaac Sim to generate contact data for a given user-defined task. We use a customizable heuristic to feed this simulation data back into the first stage to create an optimized heat map with high density in contact-rich regions. Lastly, the optimized sensor layout is used to produce a 3D printable tactile skin with individually addressable contact points.

We demonstrate the efficacy of our approach by generating six unique sensing units to cover a Franka Research 3 (FR3) robot arm and testing them in the real world for a pHRI task. 
We also generate full-surface skin designs for the Unitree humanoid H1 and quadruped Go2 to highlight the breadth of our approach. Our experimental results reveal that procedural generation is a capable tool for the design of future whole-body tactile skins. The primary contributions of this paper are: (i) a procedural generation approach to produce tactile sensor designs; 
(ii) an open-source context-driven design pipeline
% \footnote{GenTact Toolbox: https://github.com/cKohl10/GenTact}.
to procedurally generate and optimize 3D printed capacitive tactile sensors; (iii) an evaluation of our pipeline for instantiating sensors for a pHRI use case scenario.
\section{Related Work}

\subsection{Design Challenges in Whole-Body Tactile Sensing}

Previous implementations of whole-body skin sensors have focused on resolving the geometric constraints of covering robots during \textit{post-fabrication}. These sensors are often rigid modular sensors \cite{zhou2023tacsuit, maiolino2013flexible, mittendorfer2012uniform, rupavatharam2023ambisense, watanabe2021self} or flexible electronics \cite{teyssier2021human, xu2024cushsense, pannen2022low} that are designed to work in tandem with additional units. These modular sensors can wrap around the surfaces of robots to cover simple and partial curvatures while also being highly reproducible. However, they require a full re-design when the applied geometry surface changes (e.g., fitting over smaller, more complex surfaces) or when needing to alter sensor density to place the robot in a different work context.
Flexible sensors can offer a tighter fit by omitting rigid components at the expense of modularity \cite{zhao2022large}. Our pipeline conforms to robot surface geometry \textit{pre-fabrication}, allowing for tighter fitting sensors and the preservation of sensor locations upon deployment. 

Whole-body tactile sensing has also been achieved with the use of sparsely distributed external sensors. Auditory and torque signals received by such sensors can be processed with machine-learning models to localize points of contact over arbitrary robot geometries \cite{wall2023passive, iskandar2024intrinsic, fan2022enabling}. Although scalable in size, the applications of intrinsic minimal sensing methods have yet to demonstrate scalability in the amounts of simultaneous or distributed contact that can be detected due to their reliance on a sparse number of sensors. Sensor distribution was designed to be independently configurable in our pipeline to support the future of both dense and sparse whole-body tactile skins.

\subsection{3D Printed Tactile Sensors}
3D printers fabricate objects by additively stacking layers of filament, instantiating physical copies of digital meshes that closely match the models geometric complexity. Conductive filaments have advanced 3D printing by supporting electronic functionality into otherwise static objects \cite{xu2023progress, wang2023recent, wang20193d}. Particularly, capacitive sensing is a scalable and cost-effective method of tactile sensing that can utilize 3D printed conductive filament for sensing grounded objects and humans \cite{bae2023computational, navarro2021proximity}. 

Procedural generation has also been used for 3D printed components in small-scale tactile sensors, but it has not been explored as a solution to removing the design complexities of whole-body tactile skins in a self-contained manner \cite{agarwal2025vision, ruppel2022elastic}. The digital sensor configurations produced with the \textsl{GenTact Toolbox} utilize both 3D printed capacitive sensing and procedural generation to produce self-contained sensing elements.

\begin{figure}
\vspace{1.5em}
    \centering
\includegraphics[width=0.47\textwidth]{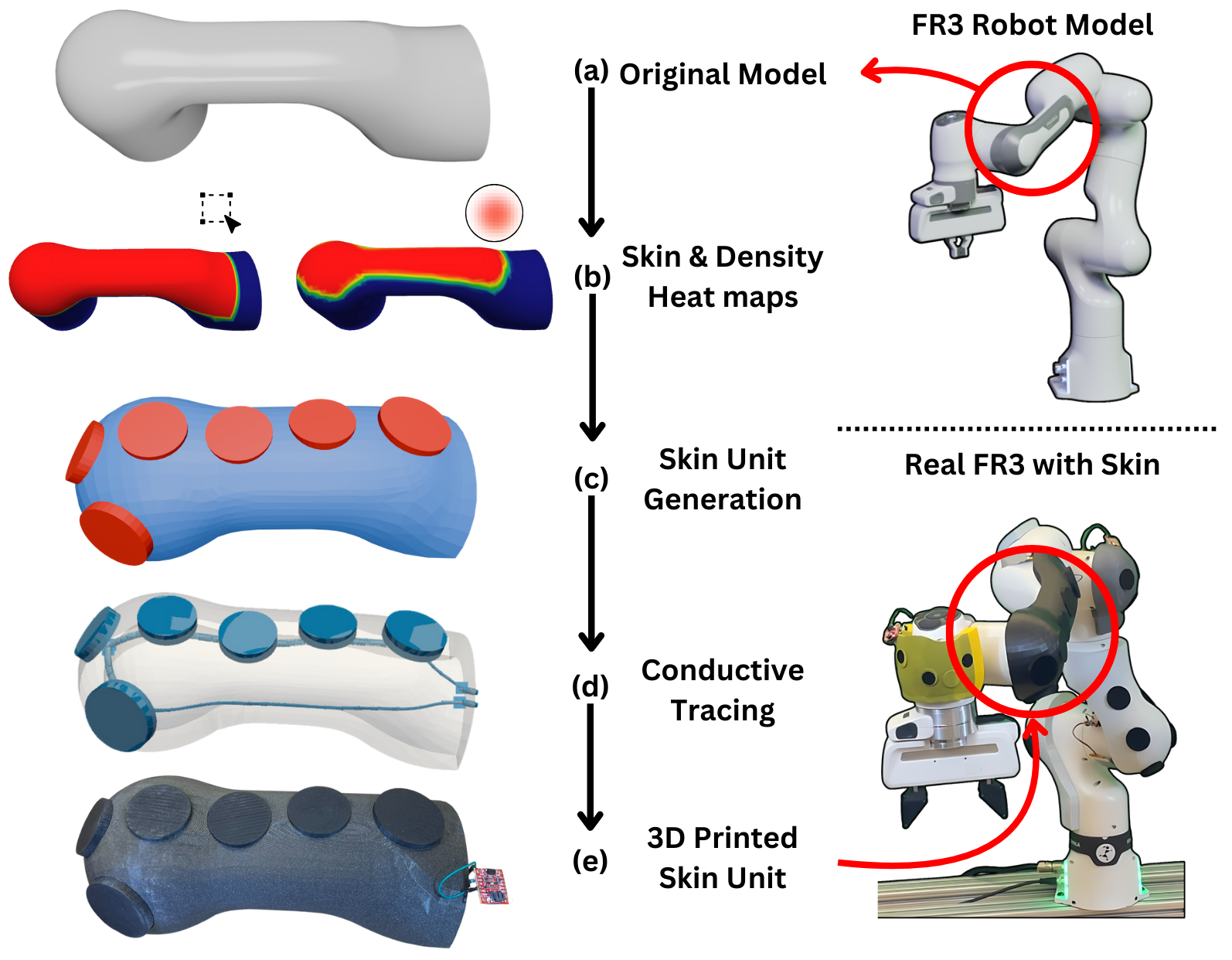}
    \caption{Snapshots of link 5 for the FR3 as it goes through the digital
 skin generation and fabrication stages of the design pipeline.}
    \label{fig:gen_example}
\end{figure}

\section{Methods}

\label{sec:design}

 The outputs of our process are modular skin coverings for individual links, referred to as skin units, that can connect wirelessly to form whole-body sensing. Skin units are designed for individual links as opposed to a robot's entirety to account for printer size limitations and to allow actuating joints to not be impeded. Each skin unit is comprised of sensing nodules---exposed conductive filament---that can measure changes in capacitance when contact occurs. Our pipeline is divided into three stages to create whole-body tactile skins that conform to the \textsl{shape} and \textsl{context} of a robot (as seen in \cref{fig:flowchart}): procedural generation, simulation, and fabrication. The code for this project is available at \url{https://github.com/HIRO-group/GenTact}

%%%%%%%%%% Section 1: Digital Skin Generation %%%%%%%%%%%%%
\subsection{Procedural Generation}
\label{sec:proced}

\label{sec2:stage1}
We procedurally generate digital skins in the open-source 3D modeling software Blender. Procedural generation refers to the algorithmic generation of content with limited or no user input \cite{viana2021procedural}. Vertices, edges, and faces of 3D models are transformed and manipulated by set geometric operations, allowing for complex meshes to be rapidly generated.

We implemented a custom add-on that leverages Blender's built-in weight painting and geometry node features to produce skins that conform to a robot's geometry (shape) and sensor distribution demands (context). A skin unit's shape is constrained by the surfaces of a given robot 3D model, assuming the real dimensions are accurately represented, to maintain a snug fit on the real robot. The skin and sensing nodule placement each use a heat map, referred to as the \textsl{skin heat map} and \textsl{density heat map} respectively, to define how they will be instantiated (as shown in \cref{fig:gen_example}-b). The heat maps are made by applying a normalized weight between zero and one to each vertex of the original model. The vertex weights can be individually addressed or painted in mass amounts with additive and subtractive digital brushes. Vertices with a skin heat map weight greater than the user-specified cutoff tolerance form the base cutout. Sharp edges are a common occurrence from taking faces directly from the original model and can be resolved with smoothing.

\begin{figure}
\vspace{1.5em}
    \centering
\includegraphics[width=0.47\textwidth]{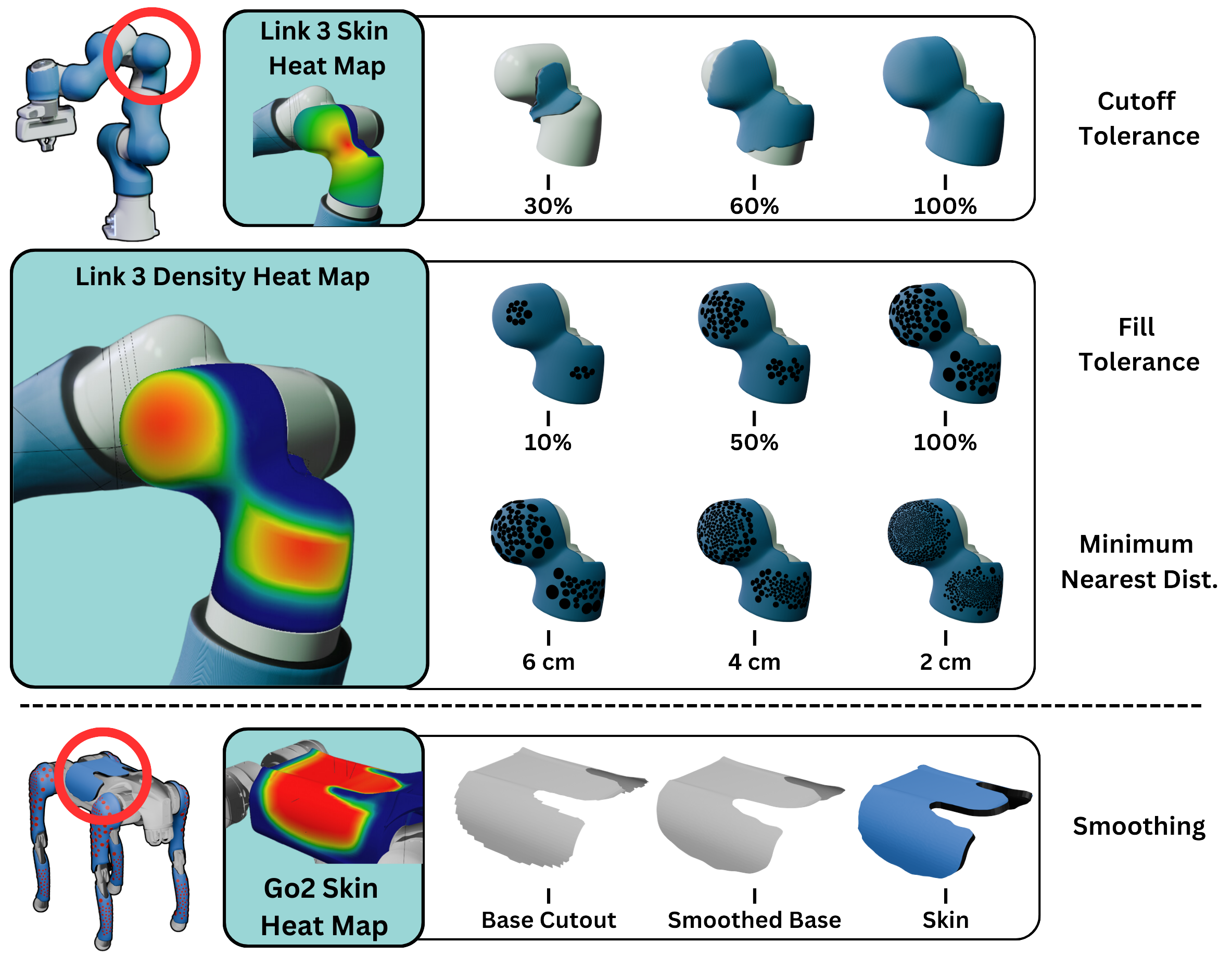}
    \caption{Top: Sensor distribution is streamlined and configurable using various scalar parameters such as the cutoff tolerance, fill tolerance, and minimum distribution distance. Bottom: The original skin heat map can produce jagged edges that require smoothing.}
    \label{fig:design_params}
\end{figure}

Sharp edges introduce risks of damaging objects and entities in the environment and thus must be removed to maintain generality towards a given robot's application. To maintain hardware compliance without compromising form, a smoothing filter is applied that gives the skin $C^1$ continuity by creating a Catmull-Rom spline around the outermost edges of the skin \cite{barsky1989geometric, twigg2003catmull}. The spline passes through the outermost vertices of the base cutout, is resampled to have less points than the original cutout, and finally reconnected to the skin by translating the original outside vertices to the nearest point along the final smoothing spline. The faces of the smoothed base cutout are then extruded outwards in their normal directions to give thickness to the skin unit for connecting sensors within a bounded volume. An example of an extruded smooth cutout is shown in blue in \cref{fig:gen_example}-c, and \cref{fig:design_params} demonstrates how the base cutout is identified and smoothed without altering the form-fitting shape of the skin unit. 

 \begin{figure*}
 \vspace{1.5em}
  \centering
  \includegraphics[width=1\linewidth]{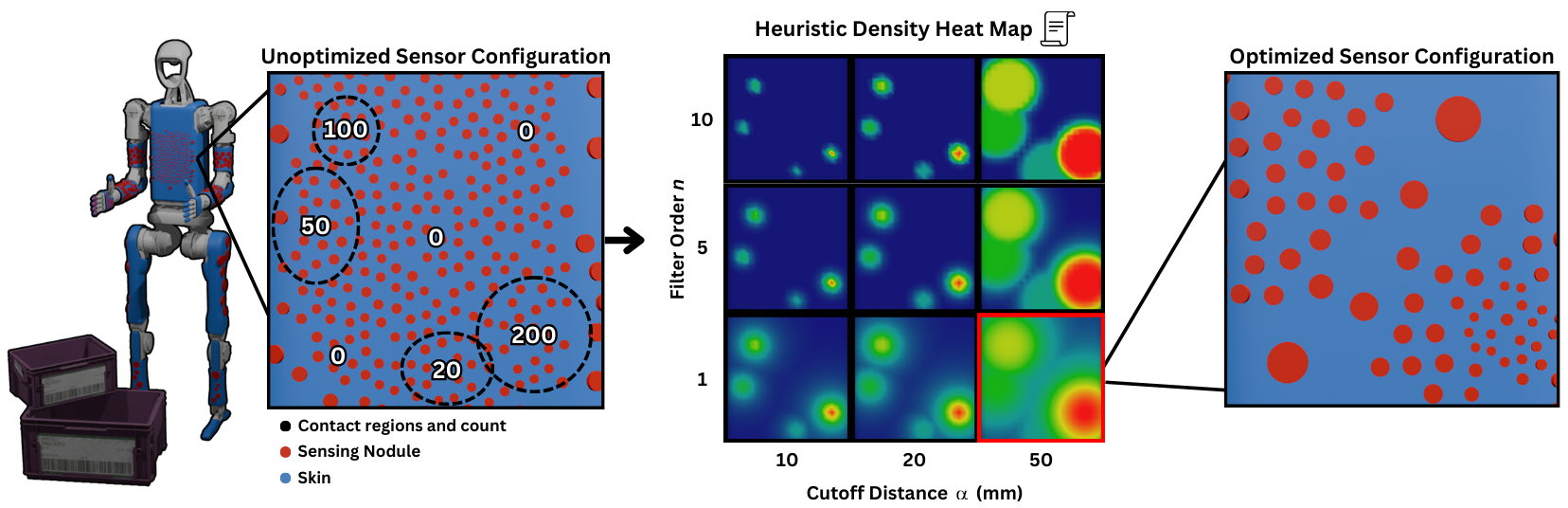}
  \caption{In this example scenario, a Unitree H1 humanoid robot is covered with unoptimized skin units and tasked to move storage bins in simulation. At the end of the simulation, four concentrated regions of high contact were identified on the chest plate. The heuristic described by \cref{eq:BW} uses the contact points to generate a new density heat map that can be tuned by $\alpha$ and $n$. The resulting optimized configuration has significantly fewer sensors in less critical regions while maintaining a high density near the more likely contact areas.}
  %\vspace{-1.5em}
  \label{fig:optimize}
\end{figure*}

The sensing nodules are placed within the skin unit and maintain a minimum distance between each other by using a Poisson-disk sampling algorithm \cite{bridson2007fast}. Poisson-disk sampling distributes nodules over the surface in a randomized fashion, eliminating those sampled under the minimum distance to another nodule until a maximum sample limit is hit. The minimum distance is set to be inversely proportional to the value of the density heat map at a given sampled location to make higher value areas in the heat map have a higher density of sensors. Varying the minimum distance provides roboticists with the design flexibility to create regions of high and low nodule density in the same skin unit. Each nodule is represented as a cylinder to make the sampled nodule locations visually distinguishable. The radius of each nodule varies in the skin unit to prevent overlap while also maximizing sensing coverage between its neighbors. A configurable radius factor between zero and one is multiplied by the minimum distance values to give designers additional flexibility in scaling the nodules.

% \begin{figure}
% \vspace{1.5em}
%     \centering
% \includegraphics[width=0.47\textwidth]{img/fr3_example_gen.png}
%     \caption{Snapshots of link 5 for the FR3 as it goes through the digital skin generation and fabrication stages of the design pipeline. The original}
%     \label{fig:fr3_example}
% \end{figure}

%%%%%%%%%% Section 2: Simulation %%%%%%%%%%%%%
\subsection{Simulation}

In order to optimize sensing nodule distribution for high densities in likely contact regions, we collect contact data in simulation and create a new density heat map using a design heuristic. The assumption made for optimizing sensor placement is that a higher density of sensors is desired in regions of frequent contact. This assumption is applicable for manipulation tasks where robots are trying to receive as much quality feedback on objects they are in contact with. We implemented a custom extension for Isaac Sim, a robotics simulation environment, to import sensor configurations, collect contact sensing data, and perform the heuristic optimization. The simulated sensing nodules are instantiated by our extension as Isaac Sim's supported PhysX contact sensors and constrained to a binary contact/no-contact output to mimic data from the real sensors discussed in \cref{sec:fabrication}.

After collecting contact data, the density heat map described in \cref{sec:proced} is recreated by a heuristic function that applies a modified Butterworth filter (\cref{eq:BW}) to each vertex in the skin unit based on the frequency and proximity of detected contact.
\begin{equation}
    v_{i,w}=\text{max}\left(v_{i,w}, \sqrt{\frac{n_{j,c}}{1+|\frac{d_{i,j}}{\alpha}|^{2n}}}\right)
    \label{eq:BW}
\end{equation}
The heuristic function starts with all weights $v_{w}$ initialized at zero and is applied to each vertex location in the skin unit $v_i$ for each nodule contact location $n_j$. In \cref{eq:BW}, $n_{j,c}$ is the number of times nodule location $n_j$ detected contact, and $d_{i,j}$ is the Euclidean distance between $v_i$ and $n_j$. The tunable parameters $\alpha$ and $n$ are the cutoff distance and filter factor of the filter respectively.
\cref{fig:optimize} showcases how an optimized heat map is formed through an example of the heuristic function applied to a Unitree H1 humanoid robot tasked with moving storage containers.

%%%%%%%%%% Section 3: Fabrication %%%%%%%%%%%%%
\subsection{Fabrication}
\label{sec:fabrication}

\begin{table*}
    \vspace{1.5em}
    \centering
    \begin{tabular}{c c c c c c | c c c c c c}
        \toprule
        \multirow{2}{*}{\textbf{Skin Unit}} & \multirow{2}{*}{\textbf{Name}} & \multirow{2}{*}{\specialcell{\hfil \textbf{Nodules}\\ }} & \multirow{2}{*}{\specialcell{\textbf{Volume}\\\textbf{(\si{\centi\meter^3})}}} & \multirow{2}{*}{\specialcell{\hfil \textbf{Total R}\\\hfil\textbf{(\si{\kilo\ohm})}}} & \multirow{2}{*}{\specialcell{\textbf{Radius}\\ \textbf{Avg.} \textbf{(\si{\centi\meter})}}} & \multirow{2}{*}{\textbf{Skin Unit}}& \multirow{2}{*}{\textbf{Name}} & \multirow{2}{*}{\specialcell{\textbf{Nodules}\\ }} & \multirow{2}{*}{\specialcell{\textbf{Volume}\\\textbf{(\si{\centi\meter^3})}}} & \multirow{2}{*}{\specialcell{\textbf{Total R}\\\textbf{(\si{\kilo\ohm})}}} & \multirow{2}{*}{\specialcell{\textbf{Radius}\\ \textbf{Avg.} \textbf{(\si{\centi\meter})}}}\\[10pt] 
        \arrayrulecolor{black!30}\midrule

    %% ------------------------------------------------------
    %% Link 1 & 4
    %% ------------------------------------------------------
    \raisebox{-\totalheight}{\includegraphics[height=0.04\textwidth]{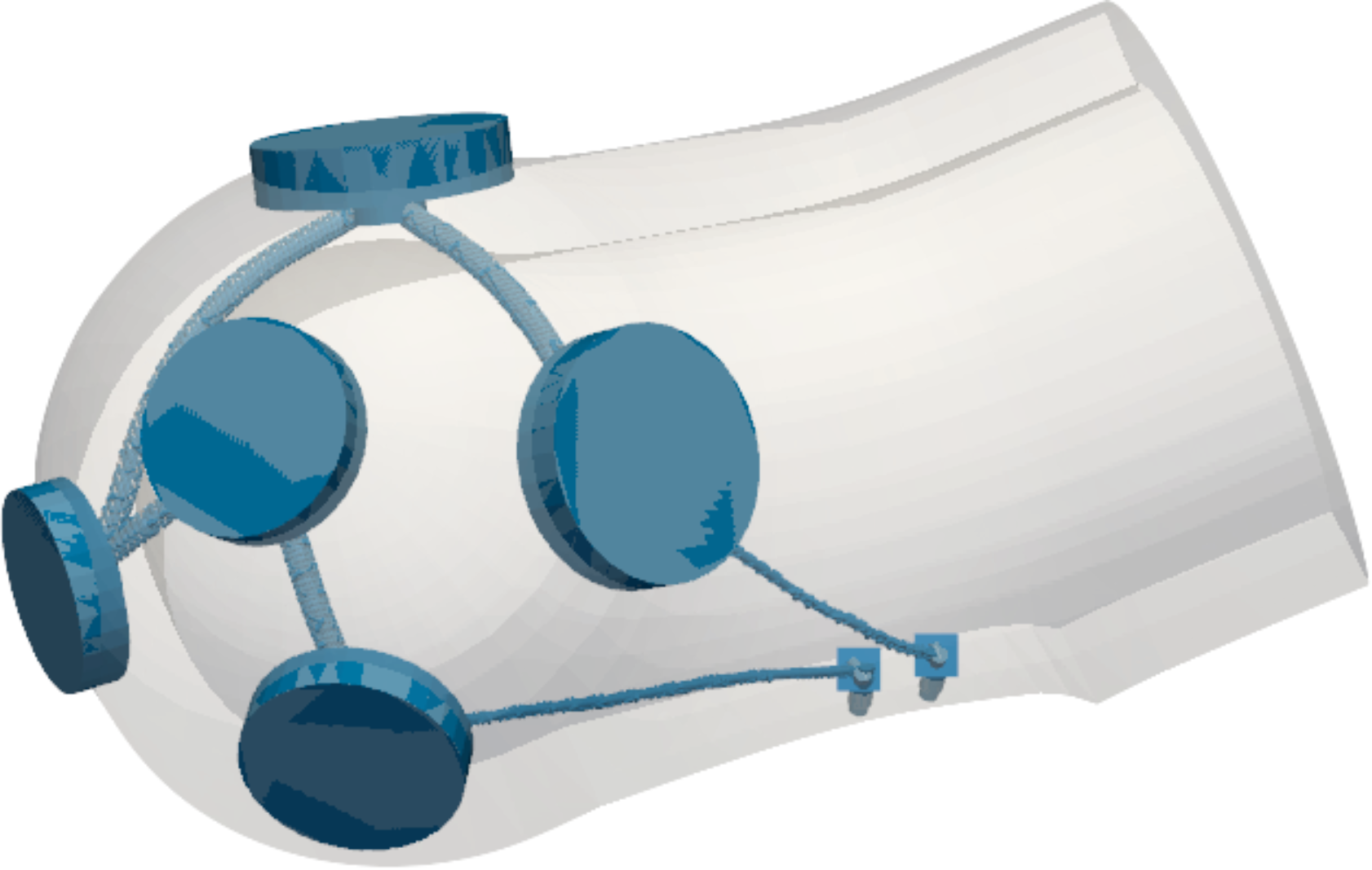}} & \multirow{4}{*}{Link 1} & \multirow{4}{*}{5} & \multirow{4}{*}{981.6} & \multirow{4}{*}{494.7} & \multirow{4}{*}{22.6} &  
    \raisebox{-\totalheight}{\includegraphics[height=0.04\textwidth]{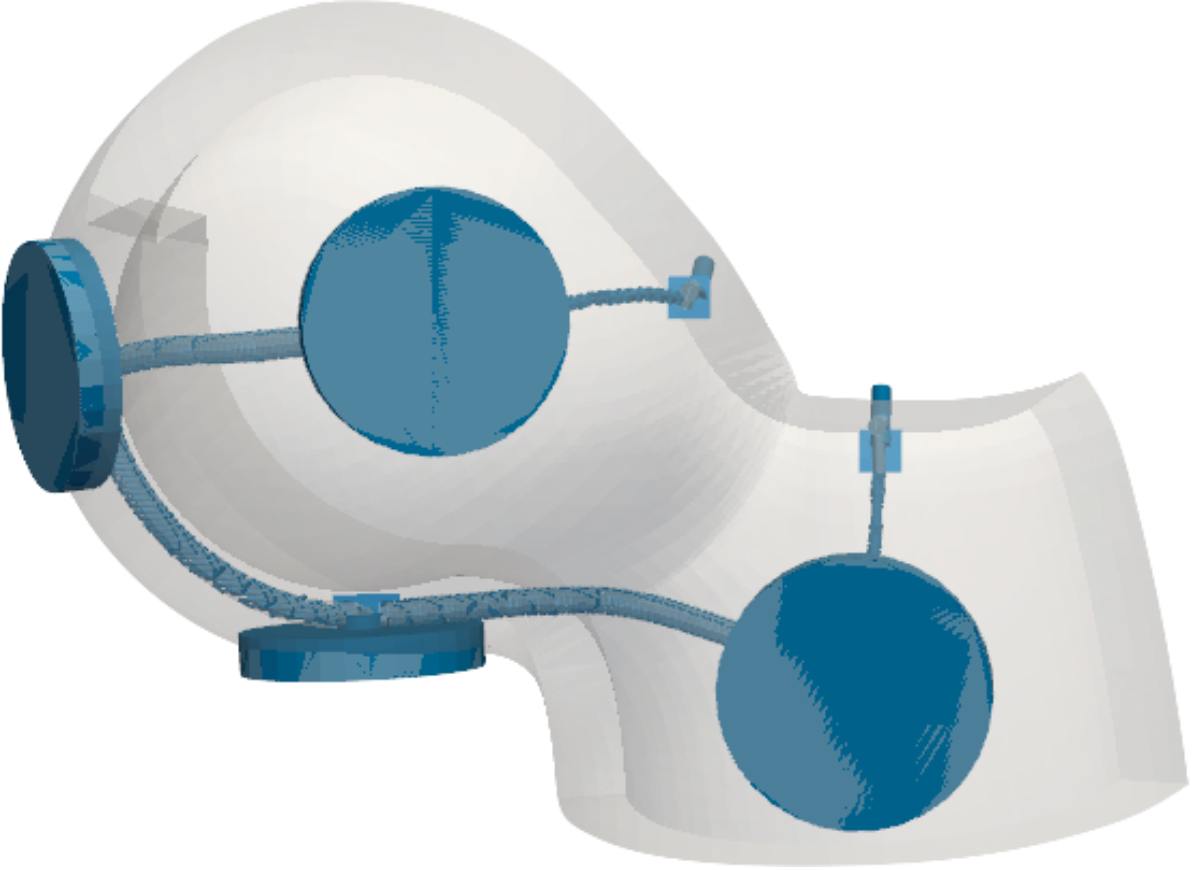}} & \multirow{4}{*}{Link 4} & \multirow{4}{*}{4} & \multirow{4}{*}{656.7} & \multirow{4}{*}{428.7} & \multirow{4}{*}{21.2} \\

    %% ------------------------------------------------------
    %% Link 2 & 5
    %% ------------------------------------------------------

    \raisebox{-\totalheight}{\includegraphics[height=0.04\textwidth]{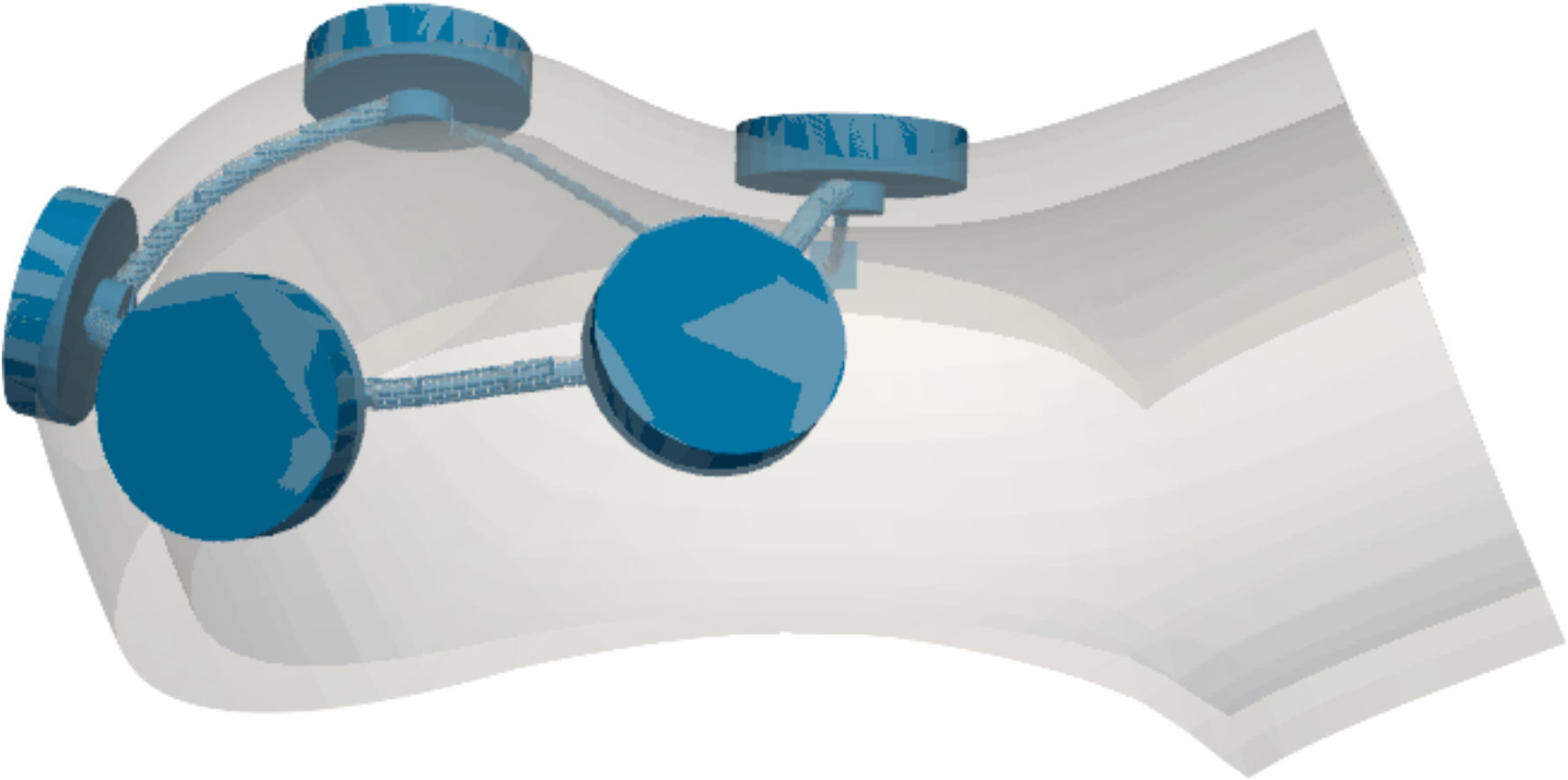}} & \multirow{4}{*}{Link 2} & \multirow{4}{*}{5} & \multirow{4}{*}{802.4} & \multirow{4}{*}{567.2} & \multirow{4}{*}{22.1} &  \raisebox{-\totalheight}{\includegraphics[height=0.035\textwidth]{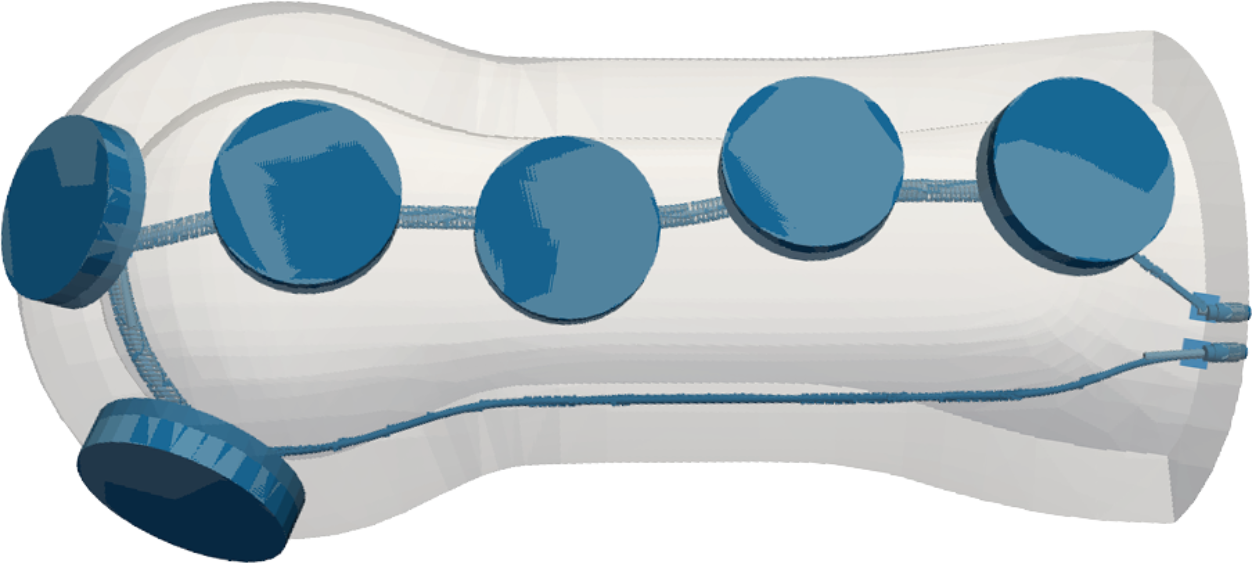}} & \multirow{4}{*}{Link 5} & \multirow{4}{*}{6} & \multirow{4}{*}{884.1} & \multirow{4}{*}{177.3} & \multirow{4}{*}{24.7}\\

    %% ------------------------------------------------------
    %% Link 3 & 6
    %% ------------------------------------------------------

    \raisebox{-\totalheight}{\includegraphics[height=0.04\textwidth]{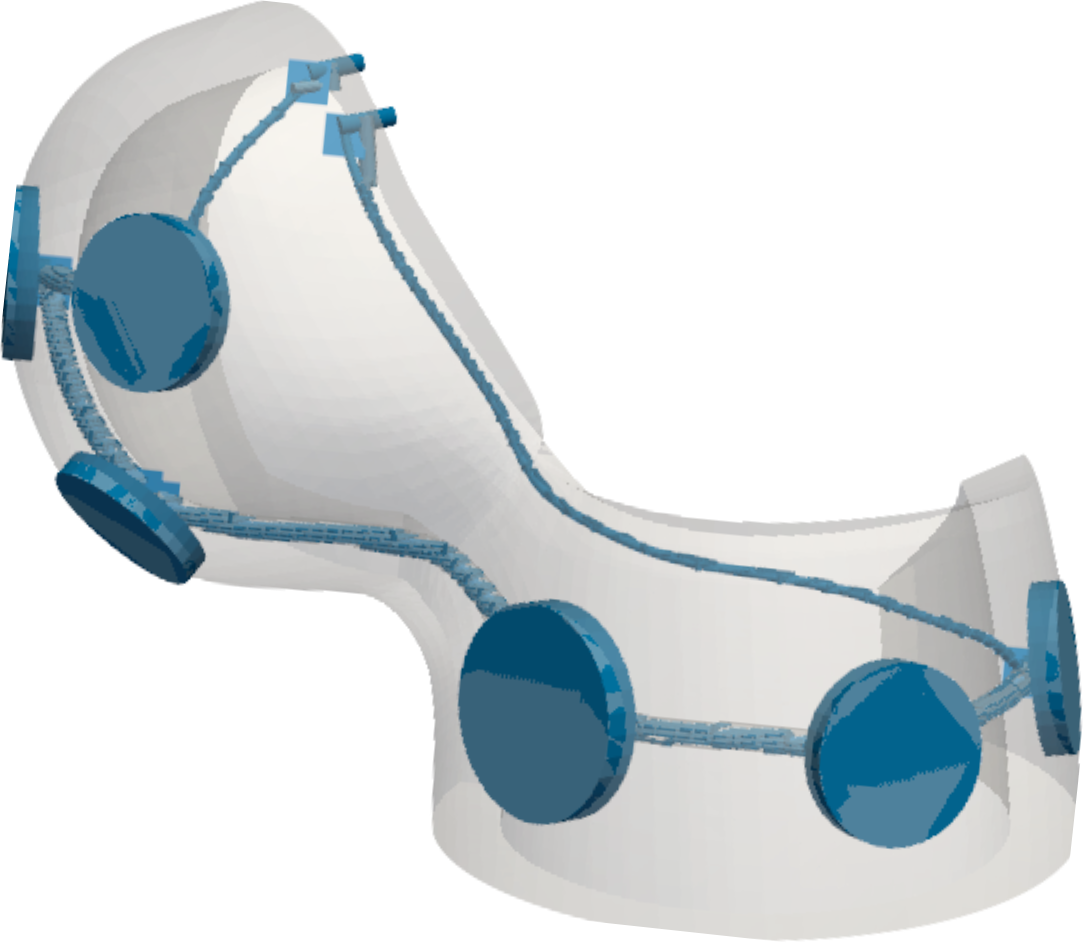}} & \multirow{4}{*}{Link 3} & \multirow{4}{*}{6} & \multirow{4}{*}{603.1} & \multirow{4}{*}{461.1} & 
    \multirow{4}{*}{20.1} & \raisebox{-\totalheight}{\includegraphics[height=0.04\textwidth]{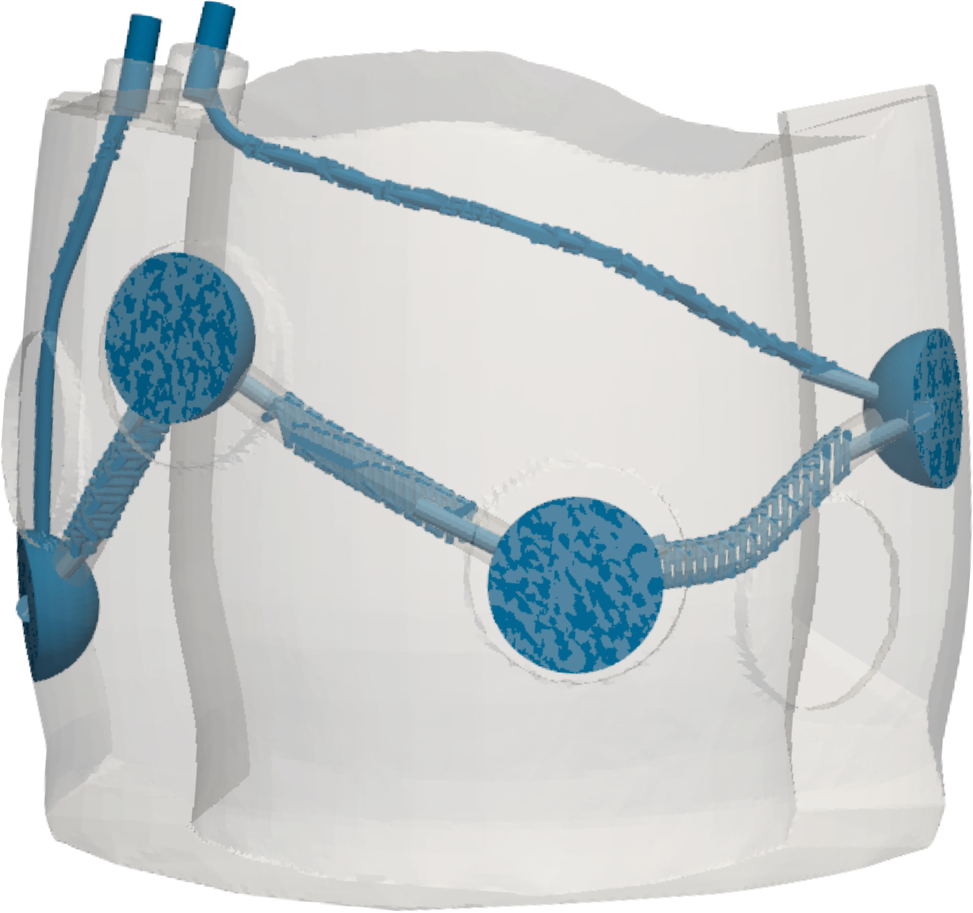}} & \multirow{4}{*}{Link 6} & \multirow{4}{*}{4} & \multirow{4}{*}{366.0} & \multirow{4}{*}{100.6} & \multirow{4}{*}{9.4}\\

    \arrayrulecolor{black}\bottomrule
    \end{tabular}
    \caption{Sensor characterizations for the six skin units for the FR3. We list each skin unit's number of nodules, mesh volume, the total resistance of the conductive traces routed throughout the unit, and the average nodule radius. The single wire connecting the nodes are shown in blue.}
    \label{tab:sensor-characteristics}
\end{table*}

\begin{table}
  \centering
  \small
  \begin{tabular}{c c c c c}
\toprule
    \textbf{Skin Unit} & \textbf{Trial 1} & \textbf{Trial 2} & \textbf{Trial 3} & \textbf{Mean}\\
\midrule
    Link 1 & 8 & 9 & 9 & 8.7 $\pm$ 0.5\\
    Link 2 & 28 & 28 & 28 & 28 $\pm$ 0\\
    Link 3 & 371 & 428 & 395 & 400 $\pm$ 20\\
    Link 4 & 33 & 35 & 33 & 34.0 $\pm$ 0.8\\
    Link 5 & 20 & 14 & 22 & 19 $\pm$ 4 \\
    Link 6 & 594 & 561 & 455 & 470 $\pm$ 70\\
\bottomrule
\end{tabular}
  \caption{Signal-to-noise ratios (SNR) for each skin unit. The minimum SNR across all nodules is reported for each trial.}
  \label{tab:snr}
\end{table}

In order to simplify the electronics fabrication and interfacing components of the \textsl{GenTact Toolbox} pipeline, we use resistor-capacitor (RC) delay sensing, which supports multiple capacitive sensing nodules with minimal instrumentation~\cite{bae2023computational}. This sensing technique optimizes the electrical resistance across the nodules such that each nodule produces a unique RC delay: the time required to charge a capacitor in a circuit through a particular amount of resistance. 
To achieve unique RC delay values, the sensing technique generates conductive traces whose geometry matches the optimized resistance values between each pair of nodules. 
These conductive traces are designed to serially connect the nodules and are printed in conductive filament enclosed by the non-conductive filament of the skin (\cref{fig:gen_example}-d). 
% Each of the nodules have a unique resistance determined by their placement in the series which causes the circuit to charge to a unique RC time delay for each nodule touched. 
After the skins are printed, the ends of the conductive trace are connected to a microcontroller (\cref{fig:gen_example}-e).
% Once there is contact with the nodules, this creates a known voltage at one end of the trace and measures the time it takes the other end to reach the same voltage within a settling threshold. 
Given how each nodule is optimized to achieve unique RC delay values via the conductive traces, the time to charge to a microcontroller's threshold voltage is distinguishable for each nodule in contact. The sensors are calibrated by manually touching each nodule and storing each measured RC delay threshold as lookup table to reference upon future contact detections.
We refer readers to Bae et al's work~\cite{bae2023computational} for more details.

\section{Evaluation}

To evaluate our pipeline, we design and characterize six skin units with the \textsl{GenTact Toolbox} to cover an FR3 tasked in a human-robot interaction use case scenario (\cref{fig:eval}). We characterize the sensing capabilities of the FR3 skin units by testing their signal qualities. While the FR3 robot thus far has served as our main example of highlighting our pipeline's capability, we evaluate the generality of our design approach by digitally generating 32 additional skin units for the Unitree Go2 quadruped robot (\cref{fig:design_params}) and Unitree H1 humanoid robot (\cref{fig:optimize}).

\subsection{Experimental Setup}
Fabrication of the sensors was done using an Original Prusa XL 5-Toolhead 3D printer with E3D ObXidian Hardened Steel nozzles and a Sunlu S4 Filament Dryer. The slicing of all models was done in PrusaSlicer (v.2.8.0). The sensing nodules were printed in Protopasta Electrically Conductive Composite PLA Filament and encased in a non-conductive Generic PLA. We used default flushing volumes and flow rates for both filaments, with hotend temperatures of \SI{230}{\celsius} for the first layer, and \SI{220}{\celsius} for the following layers of both the conductive and non-conductive filaments. The conductive filament had a resistivity per length unit of \si{256\Omega/\milli\meter} which restricted the sensing nodules to be placed with a minimum distance of \SI{9}{\milli\meter}  from the preceding nodule design pipeline.

To collect and transmit data from the sensors we used ESP32-C6 Pocket Boards soldered directly to the connection points of the sensors by melting the metal leads into the conductive plastic clips. The boards were each powered with a 
\SI{440}{\milli\ampere\hour} 1-Cell \SI{3.3}{\volt} battery. The data from the sensors were transmitted immediately after acquisition at a dynamic rate of approximately \SI{100}{\hertz}.

\subsection{Sensor Characterization}
\label{sec:characterization}
We characterized the six skin units fabricated for the FR3 by nodules, volume, trace resistance, average nodule radius, and signal-to-noise ratio (SNR) to evaluate the signal quality of skin units designed in our pipeline. SNR is a quality metric that measures signal strength against noise for inferring the likelihood of a false touch selection and informing the robustness of the sensor output (i.e., active signal) compared to disturbances of background noise (i.e., inactive signal). Following the same approach as prior work~\cite{palma2024capacitive,pourjafarian2019multi}, we measured SNR by repeatedly touching the skin units with a finger. 
For each sensing nodule, we adhered to a three-part process that lasted for 9 seconds in total. First, we did not touch for 3 seconds, then we touched the designed sensing nodule for 3 seconds and lastly let go for 3 seconds. This process was repeated for 3 trials. During this process, we measured the raw capacitive values using an Arduino Uno R4 microcontroller using the RC delay signal processing library ~\cite{bae2023computational}.

We use the following formula to compute the SNR of each skin unit: 
\begin{equation}
    \label{eq:snr}
    SNR =\frac{\left|\mu_U-\mu_P\right|}{\sigma_U}
\end{equation}
where $\mu_U$ is the mean value when the sensing nodule is not pressed, $\mu_P$ is the mean value when the sensing nodule is pressed, and $\sigma_U$ is the standard deviation of values when the sensing nodule is not pressed ~\cite{davison2010techniques}. 

Our sensors were calibrated to the individual RC delays of each nodule in contact to localize touch. The system can wrongly judge a sensing nodule selection if there is insufficient distinction in the RC delays between nodules. Hence, for our SNR calculations, we computed a pairwise calculation ($n \times n$ matrix) between the target sensing nodule ($\mu_P$) and all of the other sensing nodules ($\mu_U, \sigma_U$). We report the \textit{minimum} SNR value from this pairwise computation to illustrate the smallest gap between a pair of nodules (\cref{tab:snr}). The SNR threshold should at a minimum be 7, but ideally at least 15 for robust sensing in real-world applications~\cite{davison2010techniques}.

All skin units met the minimum SNR threshold to detect and classify the correct sensing nodules in contact, however Link 1 (Trials 1-3) and Link 5 (Trial 2) failed to meet the robustness threshold to continue making reliable detections in real-world applications without frequent calibrations. Volume appears to have an inversely proportional relationship to SNR, however the cause of the observed correlation remains inconclusive. Capacitive sensing can be impacted by changes in the operational environment. Thus, additional testing would be needed to quantify the effects of fabrication inconsistencies, parasitic capacitance, and sensor density on the signal quality.  

% \begin{figure*}
%     \vspace{1.5em}
%   \centering
%   \includegraphics[width=1\linewidth]{img/evaluation.png}
%   \caption{a) Skin units designed and fabricated for six unique links of the FR3 robot arm. A total of 30 sensing points were distributed over the surface of the arm. The skin units were then connected to an online path planner in Isaac Sim to evaluate the real-world deployment of the sensors. b) The unobstructed path of the robot moved the end-effector from left to right freely. c) In the obstructed path, contact was detected and localized, allowing the planner to generate a new path that does not intersect with the obstruction.}
%   %\vspace{-1.5em}
%   \label{fig:eval}
% \end{figure*}

% \begin{figure}
% \vspace{1.5em}
%     \centering
% \includegraphics[width=0.47\textwidth]{img/many_skins_4.png}
%     \caption{The procedural generation step in the pipeline conforms to the geometry of the robot prior to fabrication. This is shown through 38 digital skin units for various sections of the FR3, Unitree Go 2, and Unitree H1 generated in the procedural generation step of our pipeline without the use of additional modeling tools. (Not shown to scale)}
%     \label{fig:many_skins}
% \end{figure}

\begin{figure}
\vspace{1.5em}
    \centering
\includegraphics[width=0.47\textwidth]{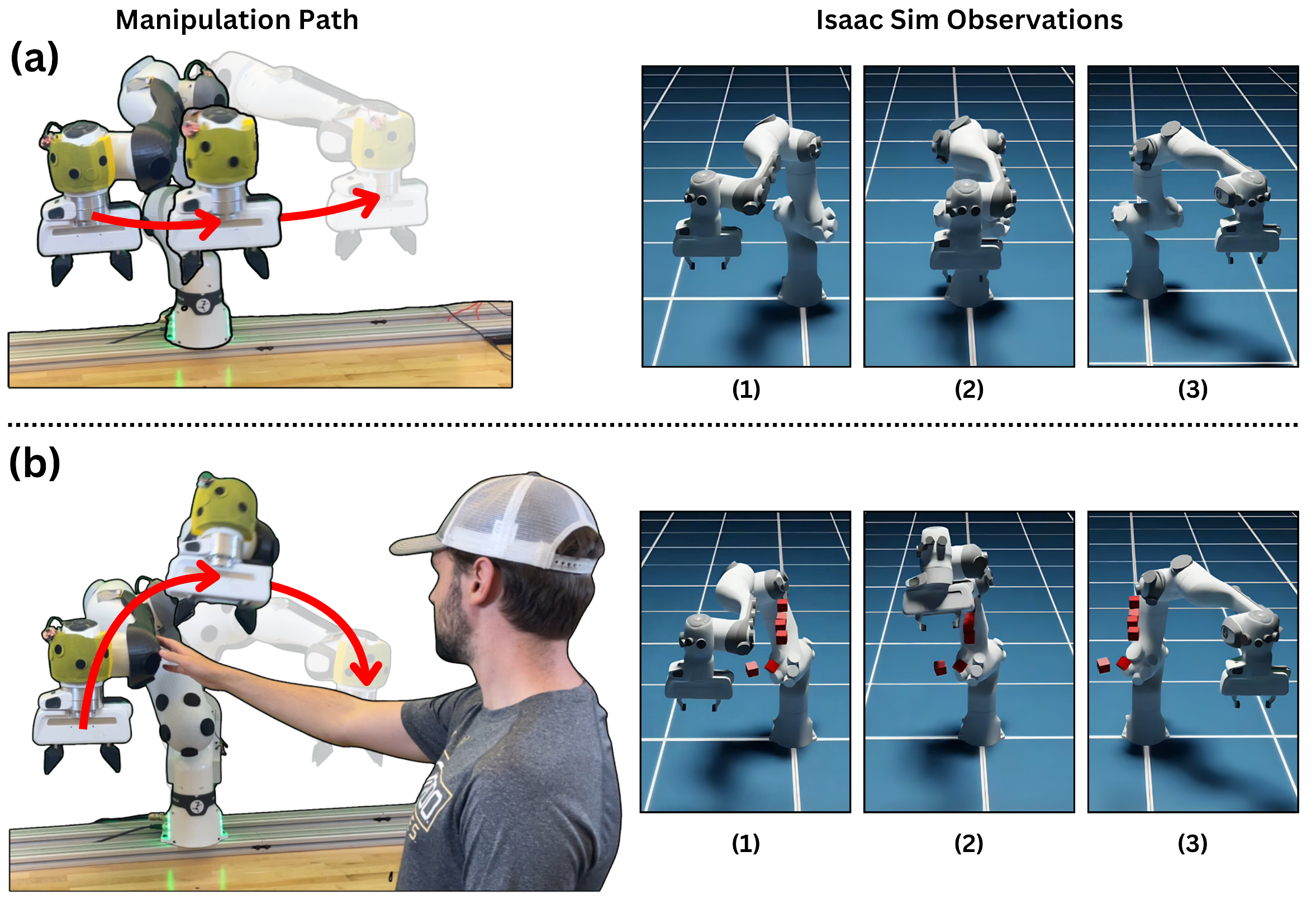}
    \caption{A total of 30 sensing points were distributed over the surface of the FR3 arm. The skin units were then connected to an online path planner in Isaac Sim to evaluate the real-world deployment of the sensors. a) The unobstructed path of the robot moved the end-effector from left to right freely. b) In the obstructed path, contact was detected and localized, allowing the planner to generate a new path that does not intersect with the obstruction.}
    \label{fig:eval}
\end{figure}

\subsection{Real-World Deployment}
To showcase the skin units created in the \textsl{GenTact Toolbox} being used in a real world scenario, we mounted the six skin units characterized in \cref{sec:characterization} to an FR3 robot and performed a pHRI task. We use the CUDA Accelerated Robot Library (cuRobo) to continuously generate collision-free trajectories between two end-effector poses \cite{curobo_report23}.
\text{\cref{fig:eval}-a} shows the unobstructed trajectory of the arm and \text{\cref{fig:eval}-b} shows an example trajectory generated upon the skin units detecting an obstruction. Detected contact was modeled as normalized voxels instantiated at the contact location. The planner was able to successfully use contact data from the skin units to avoid obstructions, exhibiting the real-world utility for safe pHRI.

\section{Conclusion and Future Work}

We presented \textsl{GenTact Toolbox}, an open-source computational pipeline that creates \textsl{form-fitting} and \textsl{context-driven} whole-body tactile skins. Our pipeline uses the algorithmic approach of procedural generation to conform tactile sensors to robot geometry. Sensors are sampled on the surface of the skin in a density determined by configurable parameters. Additionally, the placement of sensors within the skin can be defined and optimized by data, such as contact history or contact likelihood. The generated skins are finally exported as fully 3D-printable sensors capable of detecting contact with grounded objects. The real-world application of our pipeline was demonstrated in a physical human-robot interaction scenario by a Franka Research 3 robot arm fully covered in procedurally generated tactile skin units. Our approach marks a step forward in designing whole-body tactile sensors generalizable to any robot and operational task.

One of the procedural generation limitations included broken meshes on heavily concave surfaces. The pipeline failed to produce printable meshes in cases where the base cutout surface was severely concave and the skin thickness was too large, causing overlap in the mesh. The workaround found for this limitation was to lower the surface thickness enough to remove overlap or remove the concave faces from the base cutout. In addition, the density of the tactile sensors was limited at the fabrication stage by the high resistivity needed to reliably distinguish nodule signals, which was in turn dependent on the specific choice of the conductive PLA filament used in our evaluation. A minimum distance of 6 cm between nodules was experimentally found to be reliable; we aim at improving this limitation in future work. 

Procedural generation was shown to be a capable and scalable tool for embedding capacitive sensors in tactile skins. The SNR results (\cref{tab:snr}) highlight how an in-depth characterization of the sensor design (digital) and the conductive filament properties (physical) is needed to further improve sensor robustness and increase the number of sensing nodules in a single skin unit.
% limitations may reveal practical avenues to increase the maximum physical density.
We implement 3D-printed capacitive sensors as an exploratory case study to demonstrate how \textsl{GenTact Toolbox} can procedurally generate whole-body tactile sensors. However, we note that the individual stages of \textsl{GenTact Toolbox} are independent of each other and should be expanded upon to support a wider array of solutions. We believe procedural generation can be expanded to support a larger selection of sensing modalities such as pressure sensors that can sense skin deformation through flexible conductive filaments \cite{kumar2019review}. Additionally, the optimization heuristic implemented was applicable for gathering high-resolution data in contact-rich areas while reducing resolution in low-contact areas. Alternative heuristics should be explored to benefit a larger variety of operational contexts.

\bibliographystyle{IEEEtran}
\bibliography{references}
\end{document}